%% file: root.tex
\documentclass[letterpaper, 10 pt, conference, hidelinks]{ieeeconf} 

\IEEEoverridecommandlockouts                              

\usepackage{multicol}
\usepackage[english]{babel}

\usepackage{subfig}
\usepackage{booktabs}
\usepackage{color}
\usepackage{footmisc}
\usepackage[bookmarks=true,hyperfootnotes=false]{hyperref}

\usepackage{multirow}

\usepackage{amsmath} 
\usepackage{amssymb}
\usepackage{amsfonts}
\usepackage{mathtools}
\usepackage{algorithm}
\usepackage{algcompatible}
\usepackage{algpseudocode}
\usepackage{varwidth}

\usepackage{cleveref}
\crefname{equation}{equation}{eq.}
\Crefname{equation}{Equation}{Eq.}
\crefrangelabelformat{equation}{(#3#1#4--#5#2#6)}
\crefmultiformat{equation}{eq. (#2#1#3}{, #2#1#3)}{#2#1#3}{#2#1#3}
\Crefmultiformat{equation}{Eq. (#2#1#3}{, #2#1#3)}{#2#1#3}{#2#1#3}

\usepackage{siunitx}
\usepackage{graphicx}
\usepackage{tikz}
\usepackage{tikz-3dplot}
\usepackage{soul}

\usepackage{siunitx}
\usepackage{pgfplots}
\pgfplotsset{compat=newest}
\newlength\figureheight
\newlength\figurewidth
  
\usepackage{balance}

\global\long\def\inner#1#2{\left<#1,#2\right>} 
\global\long\def\bd{\mathbf{d}} 
\global\long\def\be{\mathbf{e}} 
\global\long\def\br{\mathbf{r}} 
\global\long\def\bx{\mathbf{x}}
\global\long\def\bz{\mathbf{z}} 
\global\long\def\rot{\mathbf{R}} 
\global\long\def\pos{\ensuremath{p}} 
\global\long\def\vpos{\mathbf{\pos}} 
\global\long\def\vel{\ensuremath{v}} 
\global\long\def\vvel{\mathbf{\vel}} 
\global\long\def\acc{\ensuremath{a}} 
\global\long\def\vacc{\mathbf{\acc}} 
\global\long\def\gravity{\ensuremath{g}} 
\global\long\def\vgravity{\mathbf{\gravity}} 
\global\long\def\gapl{\ensuremath{l}} 
\global\long\def\gapd{\ensuremath{d}} 
\global\long\def\t{\ensuremath{t}} 

\begin{document}	

\title{Aggressive Quadrotor Flight through Narrow Gaps \\with Onboard Sensing and Computing using Active Vision}

\author{Davide Falanga, Elias Mueggler, Matthias Faessler and Davide Scaramuzza%
\thanks{The authors are with the Robotics and Perception Group, University of Zurich, Switzerland---\url{http://rpg.ifi.uzh.ch}.
This research was funded by the DARPA FLA Program, the National Center of Competence in Research (NCCR) Robotics through the Swiss National Science Foundation and the SNSF-ERC Starting Grant.}%
}

\maketitle
\thispagestyle{empty}
\pagestyle{empty}

\input{chapters/abstract}
\input{chapters/introduction}
\input{chapters/trajectory_planning}
\input{chapters/state_estimation}
\input{chapters/experiments}
\input{chapters/discussion}

\input{chapters/conclusion}

\balance
\bibliographystyle{IEEEtran}
\bibliography{root.bbl}

\end{document}

%% file: chapters/abstract.tex
\begin{abstract}
We address one of the main challenges towards autonomous quadrotor flight in complex environments, which is flight through narrow gaps.
While previous works relied on off-board localization systems or on accurate prior knowledge of the gap position and orientation in the world reference frame,
we rely solely on onboard sensing and computing and estimate the full state by fusing gap detection from a single onboard camera with an IMU.
This problem is challenging for two reasons: (i) the quadrotor pose uncertainty with respect to the gap increases quadratically with the distance from the gap;
(ii) the quadrotor has to actively control its orientation towards the gap to enable state estimation (i.e., active vision).
We solve this problem by generating a trajectory that considers geometric, dynamic, and perception constraints:	
during the approach maneuver, the quadrotor always faces the gap to allow state estimation, while respecting the vehicle dynamics;
during the traverse through the gap, the distance of the quadrotor to the edges of the gap is maximized.
Furthermore, we replan the trajectory during its execution to cope with the varying uncertainty of the state estimate.
We successfully evaluate and demonstrate the proposed approach in many real experiments, achieving a success rate of $80\%$ and gap orientations up to \SI{45}{\degree}.
To the best of our knowledge, this is the first work that addresses and achieves autonomous, aggressive flight through narrow gaps using only onboard sensing and computing and without prior knowledge of the pose of the gap.
\end{abstract}

%% file: chapters/introduction.tex
\section*{Supplementary Material} \label{sec:additional_material}
The accompanying video is available at:\\ 
{\small \url{http://rpg.ifi.uzh.ch/aggressive_flight.html}}

\section{Introduction} \label{sec:introduction}

Recent works have demonstrated that micro quadrotors are extremely agile and versatile vehicles, able to execute very complex maneuvers~\cite{Mueller11iros,Cutler15dsmc,Mellinger10iser}.
These demonstrations highlight that one day quadrotors could be used in search and rescue applications, such as in the aftermath of an earthquake, to navigate through buildings, by entering and exiting through narrow gaps, and to quickly localize victims.

\begin{figure}[th!]
  \centering
  \begin{tabular}{cc}
  \subfloat[][The quadrotor passing through the gap.]{\includegraphics[width=0.9\linewidth]{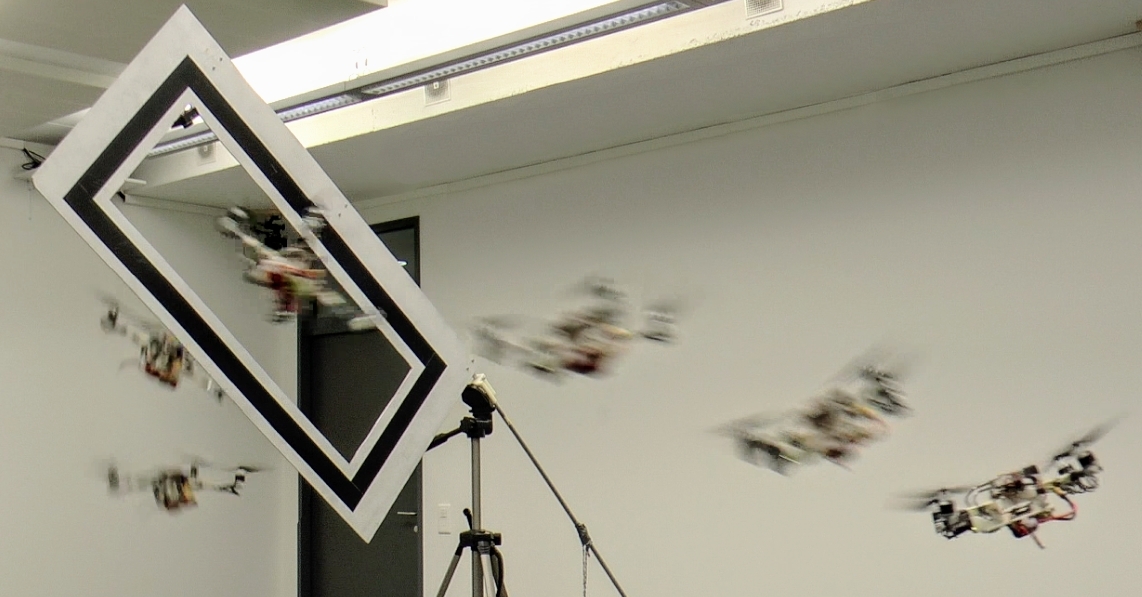}} \\
  \subfloat[][View from the onboard camera]{\includegraphics[width=0.9\linewidth]{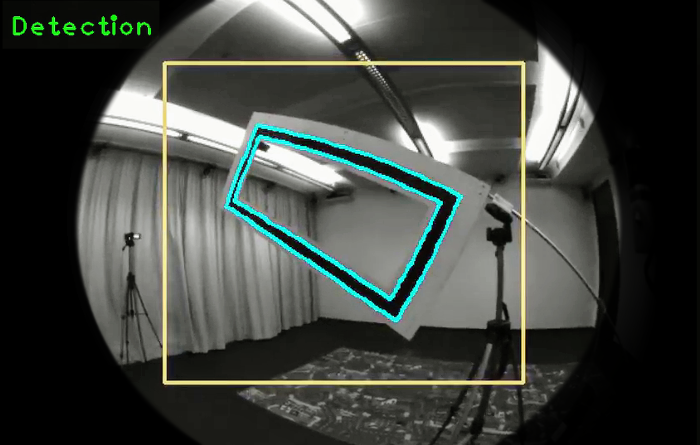}}
  \end{tabular}
  \caption{Sequence of our quadrotor passing through a narrow, \SI{45}{\degree}-inclined gap. Our state estimation fuses gap detection from a single onboard forward-facing camera with an IMU. 
					 All planning, sensing, control run fully onboard on a smartphone computer.}
  \label{fig:overview}
\end{figure}

In this paper, we address one of the main challenges towards autonomous quadrotor flight in complex environments, which is flight through narrow gaps.
What makes this problem challenging is that the gap is very small, such that precise trajectory-following is required, and can be oriented arbitrarily, such that the quadrotor cannot fly through it in near-hover conditions. 
This makes it necessary to execute an aggressive trajectory (i.e., with high velocity and angular accelerations) in order to align the vehicle to the gap orientation (cf. Fig.~\ref{fig:overview}).

Previous works on aggressive flight through narrow gaps have focused solely on the control and planning problem and therefore relied on accurate state estimation from external motion-capture systems and/or accurate knowledge of the gap position and orientation in the world reference frame. Since these systems were not gap-aware, the trajectory was generated before execution and never replanned. Therefore, errors in the measure of the pose of the gap in the world frame were not taken into account, which may lead to a collision
with gap.
Conversely, we are interested in using only \emph{onboard sensing and computing}, \emph{without any prior knowledge} of the gap pose in the world frame.
More specifically, we address the case where state estimation is done by fusing gap detection through a single, forward-facing camera with an IMU. We show that this raises an interesting \emph{active-vision} problem (i.e, coupled perception and control). 
Indeed, for the robot to localize with respect to the gap, a trajectory that guarantees that the quadrotor always faces the gap must be selected (perception constraint).
Additionally, it must be replanned multiple times during its execution to cope with the varying uncertainty of the state estimate, which is quadratic with the distance from the gap.
Furthermore, during the traverse, the quadrotor must maximize the distance from the edges of the gap (geometric constraint) to avoid collisions. 
At the same time, it must do so without relying on any visual feedback (when the robot is very close to the gap, it exits from the field of view of the camera). 
Finally, the trajectory must be feasible with respect to the dynamic constraints of the vehicle.

Our proposed trajectory generation approach is independent of the gap-detection algorithm being used; thus, to simplify the perception task, we use a gap with a black-and-white rectangular pattern (cf. Fig.~\ref{fig:overview}) for evaluation and demonstration.

\subsection{Related Work}
A solution for trajectory planning and control for aggressive quadrotor flight was presented in~\cite{Mellinger10iser}. 
The authors demonstrated their results with aggressive flight through a narrow gap, and by perching on inclined surfaces.
The quadrotor state was obtained using a motion-capture system.
To fly through a narrow gap, the vehicle started by hovering in a pre-computed position, flew a straight line towards a launch point, and then controlled its orientation to align with the gap.
The method was not {\it plug-and-play} since it needed training through \emph{iterative learning} in order to refine the launch position and velocity.
This was due to the instantaneous changes in velocity caused by the choice of a straight line for the approach trajectory.
Unlike their method, we use a technique that computes polynomial trajectories which are guaranteed to be feasible with respect to the control inputs.
The result is a smooth trajectory, compatible with the quadrotor dynamic constraints, which makes learning unnecessary.
Indeed, in realistic scenarios, such as search-and-rescue missions, we cannot afford training but must pass on the first attempt.

In~\cite{Mellinger11icra}, the same authors introduced a method to compute trajectories for a quadrotor solving a Quadratic Program, which minimizes the snap (i.e., the fourth derivative of position). 
In their experiments, agile maneuvers, such as passing through a hula-hoop thrown by hand in the air, were demonstrated using state estimation from a motion-capture system.

In~\cite{Tang15icra}, a technique that lets a quadrotor pass through a narrow gap while carrying a cable-suspended payload was presented and was experimentally validated using a motion-capture system for state estimation. 

In~\cite{Neunert16icra}, the authors proposed an unconstrained nonlinear model predictive control algorithm in which trajectory generation and tracking are treated as a single, unified problem.
The proposed method was validated in a number of experiments, including a rotorcraft passing through an inclined gap.
Like the previous systems, they used a motion-capture system for state estimation.

In~\cite{Lyu15CCECE}, the authors proposed a vision-based method for autonomous flight through narrow gaps by fusing data from a downward and a forward-looking camera, and an IMU. Trajectory planning was executed on an external computer. 
However, the authors only considered the case of an horizontal gap, therefore no agile maneuver was necessary.

In~\cite{Loianno16RAL}, the authors proposed methods for onboard vision-based state estimation, planning, and control for small quadrotors, and validated the approach in a number of agile maneuvers, among which flying through an inclined gap.
Since state estimation was performed by fusing input from a downward-looking camera and an IMU, rather than from gap detection, the gap position and orientation in the world reference frame had to be measured very accurately prior to the execution of the maneuver. The trajectory was generated before execution and never replanned.
Therefore, errors in the measure of the pose of the gap in the world frame were not taken into account, which may lead to a collision with gap. To deal with this issue, the authors used a gap considerably larger than the vehicle size.

All the related works previously mentioned relied on the accurate state estimates from a motion-capture system or accurate prior knowledge of the gap position and orientation in the world reference frame.
Additionally, in all these works but~\cite{Neunert16icra} and~\cite{Loianno16RAL} trajectory generation was performed on an external computer.
The advantages of a motion-capture system over onboard vision are that the state estimate is always available, at high frequency, accurate to the millimeter, and with almost {\it constant noise covariance} within the tracking volume.
Conversely, a state estimate from onboard vision can be intermittent (e.g., due to misdetections); furthermore, its covariance increases \emph{quadratically} with the distance from the scene and is strongly affected by the type of structure and texture of the scene.
Therefore, to execute a complex aggressive maneuver, like the one tackled in this paper, while using only onboard sensing and \emph{gap-aware} state estimation, it becomes necessary \emph{to couple perception with the trajectory generation process} (i.e., active vision).
Specifically, the desired trajectory has to render the gap always visible by the onboard camera in order to estimate its relative pose.

\subsection{Contributions}

Our method differs from previous works in the following aspects: 
(i)~we rely solely on onboard, visual-inertial sensors and computing,
(ii)~we generate a trajectory that facilitates the perception task, while satisfying geometric and dynamic constraints, and
(iii)~we do not require iterative learning, neither do we need to know a priori the gap position and orientation in the world frame.
To the best of our knowledge, this is the first work that addresses and achieves aggressive flight through narrow gaps with state estimation via gap detection from an onboard camera and IMU.

The remainder of this paper is organized as follows. 
Section~\ref{sec:planning} presents the proposed trajectory-generation algorithm.
Section~\ref{sec:state_estimation} describes the state-estimation pipeline. 
Section~\ref{sec:experiments} presents the experimental results.
Section~\ref{sec:discussion} discusses the results and provides additional insights about the approach. 
Finally, Section~\ref{sec:conclusion} draws the conclusions.

%% file: chapters/trajectory_planning.tex
\section{Trajectory Planning} \label{sec:planning}
We split the trajectory planning into two consecutive stages. 
First, we compute a traverse trajectory to pass through the gap. 
This trajectory maximizes the distance from the vehicle to the edges of the gap in order to minimize the risk of collision.
In a second stage, we compute an approach trajectory in order to fly the quadrotor from its current hovering position to the desired state that is required to initiate the traverse trajectory.
While both trajectories need to satisfy \emph{dynamic constraints}, the approach trajectory also satisfies \emph{perception constraints}, i.e., it lets the vehicle-mounted camera always face the gap.
This is necessary to enable state estimation with respect to the gap.
 
\subsection{Traverse Trajectory} \label{sec:ballistic_trajectory}
\begin{figure}[!htb]
\centering
\includegraphics[width=0.5\linewidth]{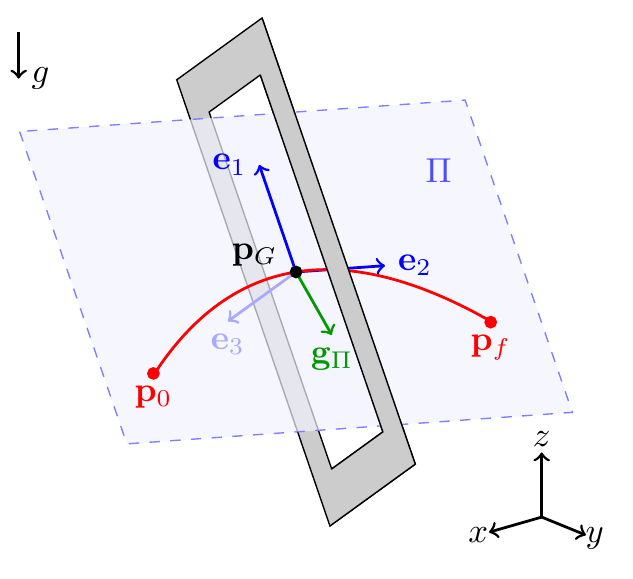}
\caption{An inclined gap and the corresponding plane $\Pi$.}
\label{fig:window_and_plane}
\end{figure}

During the gap traversal, the quadrotor has to minimize the risk of collision.
We achieve this by forcing the traverse trajectory to intersect the center of the gap while simultaneously lying in a plane orthogonal to the gap (see Fig.~\ref{fig:window_and_plane}).
In the following, we derive the traverse trajectory in this orthogonal plane and then transform it to the 3D space.

Let~${W}$ be our world frame. 
The vector $\vpos_G$ and the rotation matrix $\rot_G$ denote the position of the geometric center of the gap and its orientation with respect to~${W}$, respectively.
Let~$\Pi$ be a plane orthogonal to the gap, passing through its center and parallel to the longest side of the gap (cf. Fig.~\ref{fig:window_and_plane}).
Let~$\be_1$ and~$\be_2$ be the unit vectors spanning such a plane~$\Pi$, whose normal unit vector is~$\be_3$. 
The $\be_2$ axis is orthogonal to the gap and $\be_1 = \be_2 \times \be_3$.

\begin{figure}[tb]
\centering
\includegraphics[width=0.5\linewidth]{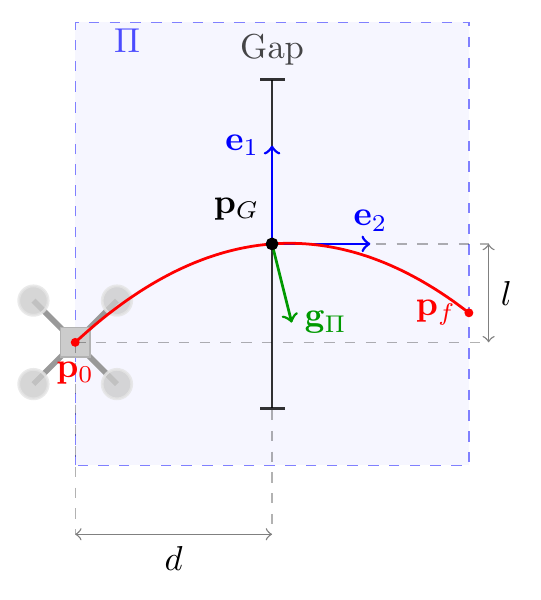}
\caption{The traverse trajectory in the plane $\Pi$.}
\label{fig:in_plane_motion}
\end{figure}

Intuitively, a trajectory that lies in the plane~$\Pi$ and passes through the center of the gap, minimizes the risk of impact with the gap.

To constrain the motion of the vehicle to the plane~$\Pi$, it is necessary to compensate the projection of the gravity vector~$\vgravity$ onto its normal vector~$\be_3$. Therefore, a constant thrust of magnitude~$\inner{\vgravity}{\be_3}$ needs to be applied orthogonally to~$\Pi$. 
By doing this, a 2D description of the quadrotor's motion in this plane is sufficient.
The remaining components of~$\vgravity$ in the plane~$\Pi$ are computed as
\begin{equation} \label{eq:gravity_in_plane}
	\vgravity_{\Pi} = \vgravity - \inner{\vgravity}{\be_3} \be_3 .
\end{equation}

Since this is a constant acceleration, the motion of the vehicle along $\Pi$ is described by the following second order polynomial equation:
\begin{subequations}\label{eq:parabola_in_plane}
\begin{align} 
	\pos_i(\t) &= \pos_i(\t_0) + \vel_i(t_0)\t + \frac{1}{2}\gravity_{\Pi,\mathrm{{i}}}\t^2, \\
	\vel_i(\t) &= \vel_i(\t_0) + \gravity_{\Pi,\mathrm{i}}\t  ,
\end{align}
\end{subequations}
where the subscript~$i=\{1, 2\}$ indicates the component along the~$\be_i$ axis. 
The quadrotor enters the traverse trajectory at time~$\t_0$, $\t$ is the current time, and~$\pos$ and~$\vel$ denote its position and velocity, respectively.

Equation~\eqref{eq:parabola_in_plane} describes a ballistic trajectory.
When $\gravity_{\Pi,\mathrm{2}}~=~0$, it is the composition of a uniformly accelerated and a uniform-velocity motion. 
In other words, in these cases the quadrotor moves on a parabola in space.

Let $\gapl$ and $\gapd$ be the distance between $\vpos_G$ and the initial point of the trajectory, $\vpos_0$, along $\be_1$ and $\be_2$, respectively (cf. Fig.~\ref{fig:in_plane_motion}).
These two parameters determine the initial position and velocity in the plane $\Pi$, 
as well as the time $\t_c$ necessary to reach $\vpos_G$. 
The values of~$\gapd$ and~$\gapl$ are determined through an optimization problem, as explained later in Sec.~\ref{sec:traverse_optimization}.

For a generic orientation $\rot_G$ of the gap, \eqref{eq:parabola_in_plane} is characterized by a uniformly accelerated motion along both the axes~$\be_1$ and~$\be_2$. 
Therefore, it is not possible to guarantee that the distance traveled along the $\be_2$ axis before and after the center of the gap are equal while also guaranteeing that the initial and final position have the same coordinate along the $\be_1$ axis. 
For safety reasons, we prefer to constrain the motion along the $\be_2$ axes, i.e., orthogonally to the gap, such that the distances traveled before and after the gap are equal. 

Given the components of the unit vectors $\be_1$ and $\be_2$ in the world frame, it is now possible to compute the initial conditions $\vpos_0 = \vpos(\t_0)$ and $\vvel_0 = \vvel(\t_0)$ in 3D space as follows: 
\begin{subequations} \label{eq:initial_conditions_in_3d}
\begin{align}
\vpos_0 &= \vpos_G -\gapl \be_1 - \gapd \be_2, \label{eq:initial_position_3d}\\
\vvel_0 &= \left (\frac{\gapl}{\t_c} - \frac{1}{2} \gravity_{\Pi,\mathrm{1}} \t_c \right) \be_1 + 
\left (\frac{\gapd}{\t_c} - \frac{1}{2} \gravity_{\Pi,\mathrm{2}} \t_c \right) \be_2, \label{eq:initial_velocity_3d}
\end{align}
\end{subequations}
where:
\begin{equation} \label{eq:time_for_the_peak}
t_c = \sqrt{\frac{-2\gapl}{\gravity_{\Pi,\mathrm{1}}}}
\end{equation}
is the time necessary to reach the center of the gap once the traverse trajectory starts.

Note that this solution holds if $\gravity_{\Pi,\mathrm{2}} \geq 0$ which applies if $\be_2$ is horizontal or pointing downwards in world coordinates.
The case $\gravity_{\Pi,\mathrm{2}} < 0$ leads to similar equations, which we omit for brevity.
The final three-dimensional trajectory then has the following form:
\begin{subequations} \label{eq:parabola_in_3d}
\begin{align}
\vpos(\t) &= \vpos_0 + \vvel_0\t + \frac{1}{2}\vgravity_{\Pi}\t^2, \label{eq:position_3d} \\ 
\vvel(\t) &= \vvel_0 + \vgravity_{\Pi}\t, \label{eq:velocity_3d}\\
\vacc(\t) &= \vgravity_{\Pi} \label{eq:acceleration_3d}.
\end{align}
\end{subequations} 

This trajectory is inexpensive to compute since it is solved in closed form.
Also, note that during the traverse the gap is no longer detectable.
Nevertheless, since the traverse trajectory is short and only requires \emph{constant control inputs} (a thrust of magnitude~$\inner{\vgravity}{\be_3}$ and zero angular velocities), it is possible to track it accurately enough to not collide with the gap, even without any visual feedback.

\subsection{Optimization of the Traverse Trajectory} \label{sec:traverse_optimization}
To safely pass through the gap, the quadrotor must reach the initial position and velocity of the traverse trajectory described by~\eqref{eq:initial_position_3d}-\eqref{eq:initial_velocity_3d} with an acceleration equal to $\vgravity_{\Pi}$  at time $\t_0$.
An error in these initial conditions is propagated through time according to~\eqref{eq:position_3d}-\eqref{eq:acceleration_3d}, and therefore may lead to a collision.
The only viable way to reduce the risk of impact is to reduce the time duration of the traverse.
More specifically, \eqref{eq:time_for_the_peak} shows that one can optimize the value of $\gapl$ to reduce the time of flight of the traverse trajectory.
On the other hand, \eqref{eq:initial_velocity_3d} and~\eqref{eq:time_for_the_peak} show that reducing $\gapl$ leads to an increase in the norm of the initial velocity $\vvel_0$.
Intuitively speaking, this is due to the fact that, for a given value of $\gapd$, if the time of flight decreases, the velocity along the $\be_2$ axis has to increase to let the vehicle cover the same distance in a shorter time.
The initial velocity also depends on $\gapd$, which can be tuned to reduce the velocity at the start of the traverse.
The value of $\gapd$ cannot be chosen arbitrarily small for two reasons: 
(i) it is necessary to guarantee a safety margin between the quadrotor and the gap at the beginning of the traverse; 
(ii) the gap might not be visible during the final part of the approach trajectory.
For this reason, we compute the values of the traverse trajectory parameters solving the following optimization problem:
\begin{equation} \label{eq:parabola_optimization_problem}
\min_{\gapd, \gapl} \t_c  \ \ \text{s.t.} \ \ \left\lVert \vvel_0 \right\rVert \leq \mathrm{v}_{\mathrm{0,max}}, \ \ \gapd \geq \gapd_{\mathrm{min}},
\end{equation}
where
$\mathrm{v}_{\mathrm{0,max}}$ and $\gapd_{\mathrm{min}}$ are the maximum velocity allowed at the start of the traverse and the minimum value of $\gapd$, respectively.
We solve the nonlinear optimization problem described by~\eqref{eq:parabola_optimization_problem} with Sequential Quadratic Programming (SQP~\cite{kraft1988sqp}, using to the NLopt library~\cite{johnsonNLOpt}.
Thanks to the small dimensionality of the problem, it can be solved onboard in few tens of milliseconds.

\subsection{Approach Trajectory} \label{sec:approach_trajectory}
Once the traverse trajectory has been computed, its initial conditions (namely, position, velocity, and acceleration) are known.
Now we can compute an approach trajectory from a suitable start position to these initial conditions.
Note that this start position is not the current hover position but also results from the proposed trajectory generation method.
Our goal in this step is to find a trajectory that not only matches the initial conditions of the traverse trajectory, but also enables robust perception and state estimation with respect to the gap.

Robust state estimation with respect to the gap can only be achieved by always keeping the gap in the field of view of a forward-facing camera onboard the quadrotor.
Since it is difficult to incorporate these constraints into the trajectory generation directly, we first compute trajectory candidates and then evaluate their suitability for the given perception task.
To do so, we use the approach proposed in~\cite{Mueller15tro}, where a fast method to generate feasible trajectories for flying robots is presented. 
In that paper, the authors provide both a closed-form solution for motion primitives that minimize the jerk and a feasibility check on the collective thrust and angular velocities.
The benefit of using such a method is twofold.
First, it allows us to obtain a wide variety of candidate trajectories within a very short amount of time by uniformly sampling the start position and the execution time within suitable ranges.
This way we can quickly evaluate a large set of candidate trajectories and select the best one according to the optimality criterion described in Sec.~\ref{sec:choice_of_trajectory}.
Each of these candidate trajectories consists of the quadrotor's 3D position and its derivatives.
Second, and most importantly, since the computation and the verification of each trajectory takes on average a two tenths of millisecond, it is possible to replan the approach trajectory at each control step, counteracting the effects of the uncertainty in the pose estimation of the quadrotor when it is far away from the gap.
Each new approach trajectory is computed using the last state estimate available.
In the following, we describe how we plan a yaw-angle trajectory for each candidate and how we select the best candidate to be executed.

\subsection{Yaw-Angle Planning} \label{sec:yaw_angle}
 
In~\cite{Mellinger11icra}, the authors proved that the dynamic model of a quadrotor is \emph{differentially flat}. 
Among other things, this means that the yaw angle of the quadrotor can be controlled independently of the position and its derivatives.
In this section, we present how to compute the yaw angle such that a camera mounted on the quadrotor always faces the gap.
Ideally, the camera should be oriented such that the center of the gap is projected as close as possible to the center of the image, which yields the maximum robustness for visual state estimation with respect to the gap against disturbances on the quadrotor.

To compute the desired yaw angle, we first need to compute the ideal orientation of the camera.
Let $\vpos_G$ be the coordinates of the center of the gap with respect to the world frame~$W$. 
Furthermore, let $\rot_{WC}$ and $\vpos_C$ be the extrinsic parameters of the camera:
$\vpos_C$ is the camera's position and the rotation matrix $\rot_{WC}=(\br_1,\br_2,\br_3)$ defines the camera orientation with respect to the world frame, where $\br_3$ is the camera's optical axis.

For a given trajectory point, we can compute the vector from the camera to the center of the gap $\bd = \vpos_G - \vpos_C$.
Ideally, we can now align the camera's optical axis $\br_3$ with $\bd$ but since the trajectory constrains the quadrotor's vertical axis $\bz_b$, we can generally not do this.
Therefore, we minimize the angle between $\bd$ and $\br_3$ by solving the following constrained optimization problem:
\begin{equation}\label{eq:optical_axis_opt_problem}
\br_3^* = \arg \max_{\bx} \inner{\bx}{\bd} \ \ \text{s.t.}  \ \| \bx \|=1, \ \ \inner{\bx}{\bz_b}=k,
\end{equation}
where the last constraint says that the angle between the quadrotor's vertical body axis $\bz_b$ and the camera's optical axis is constant and depends on how the camera is mounted on the vehicle.
For example, $k=0$ if the camera is orthogonal to the $\bz_b$ axis as it is the case in our setup with a forward-facing camera.

Letting $\bd_{\perp z_{b}} = \bd - \inner{\bd}{\bz_b}\bz_b$ be the component of $\bd$ perpendicular to $\bz_b$, the solution of~\eqref{eq:optical_axis_opt_problem} is
\begin{equation}\label{eq:optical_axis_opt_problem_solution}
\br_3^* = \sqrt{1-k^2} \frac{\bd_{\perp z_{b}}}{\|\bd_{\perp z_{b}}\|} + k \bz_b,
\end{equation}
which is a vector lying in the plane spanned by $\bd$ and $\bz_b$, 
and the minimum angle between the ideal and the desired optical axis is $\arccos (\inner{\br_3^*}{\bd}/\|\bd\|)$, i.e.,
\begin{equation}\label{eq:minAngleOpticalAxes}
\theta_{min} = 
\arccos \bigl((\sqrt{1-k^2}\|\bd_{\perp z_{b}}\| + k\inner{\bd}{\bz_b})\, /\, \|\bd\|\bigr).
\end{equation}
Once $\br_3^*$ is known, we can compute the yaw angle such that the actual camera optical axis $\br_3$ is aligned with~$\br_3^*$.

Observe that in the particular case of a trajectory point that allows to align $\br_3$ with $\bd$, we have $\inner{\bd}{\bz_b}=k\|\bd\|$ and the solution of~\eqref{eq:optical_axis_opt_problem} reduces to $\br_3^* = \frac{\bd}{\|\bd\|}$, with a minimum angle $\theta_{\mathrm{min}} = \arccos (\inner{\br_3}{\bd}/\|\bd\|) = \arccos(1) = 0$.

\subsection{Selection of the Approach Trajectory to Execute} \label{sec:choice_of_trajectory}

In the previous sections, we described how we compute a set of candidate trajectories in 3D space and yaw for approaching the gap.
All the candidate trajectories differ in their start position and their execution time.
From all the computed candidates, we select the one that provides the most reliable state estimate with respect to the gap.
As a quality criterion for this, we define a cost function $J$ composed of two terms:
\begin{itemize}
\item the Root Mean Square (RMS) $\theta_{\mathrm{rms}}$ of~\eqref{eq:minAngleOpticalAxes} over every sample along a candidate trajectory;
\item the straight-line distance $d_{\textrm{0}}$ to the gap at the start of the approach.
\end{itemize}
More specifically:
\begin{equation} \label{eq:approach_cost_function}
J = \frac{\theta_{\mathrm{rms}}}{\bar{\theta}} + \frac{d_{\textrm{0}}}{\bar{d}},
\end{equation}
where $\bar{\theta}$ and $\bar{d}$ are normalization constants that make it possible to sum up quantities with different units, and render the cost function dimensionless.
This way, the quadrotor executes the candidate approach trajectory that keeps the center of the gap as close as possible to the center of the image for the entire trajectory, and at the same time prevents the vehicle from starting too far away from the gap.

\subsection{Recovery after the Gap} \label{sec:recovery}

Since we localize the quadrotor with respect to the gap in order to traverse it, the quadrotor is left with no state estimate after the traversal.
Therefore, at this point it has to recover a vision-based state estimate and then hover in a fixed position without colliding with the environment.
We solve this problem using the automatic recovery system detailed in~\cite{Faessler15icra}, where the authors provide a method to let a quadrotor stabilize automatically after an aggressive maneuver, e.g. after a manual throw in the air.

%% file: chapters/state_estimation.tex
\section{State Estimation} \label{sec:state_estimation}

\subsection{State Estimation from Gap Detection} \label{sec:window_tracking}
Our proposed trajectory generation approach is independent of the gap-detection algorithm being used; 
thus, to simplify the perception task, we use a black-and-white rectangular pattern to detect the gap (cf.~Fig.~\ref{fig:overview}).
A valid alternative to cope with real-world gaps would be to use monocular dense-reconstruction methods, such as REMODE~\cite{Pizzoli14icra}; however, they require more computing power (GPUs).

We detect the gap in each image from the forward-facing camera by applying a sequence of steps:
first, we run the Canny edge detector, undistort all edges, and group close edges~\cite{Suzuki85cvgip};
then, we search for quadrangular shapes and run geometrical consistency checks.
Namely, we search for a quadrangle that contains another one and check the area ratio of these two quadrangles.
Finally, we refine the locations of the eight corners to sub-pixel accuracy using line intersection. 

Since the metric size of the gap is known, we estimate the 6-DOF pose by solving a Perspective-n-Points (PnP) problem (where $n=8$ in our case).
As a verification step, we require that the reprojection error is small.
We then refine the pose by minimizing also the reprojection error of all edge pixels.
To speed up the computation, we only search the gap in a region of interest around the last detection. 
Only when no detection is found, the entire image is searched.
The detector runs with a frequency of more than \SI{30}{\hertz} onboard the quadrotor.

Finally, we fuse the obtained pose with IMU measurements to provide a full state estimate using the multi-sensor fusion framework of~\cite{Lynen13iros}.

%% file: chapters/experiments.tex
\section{Experiments} \label{sec:experiments}

\subsection{Experimental Setup}
\begin{figure}[tb]
  \centering
  \includegraphics[width=\linewidth]{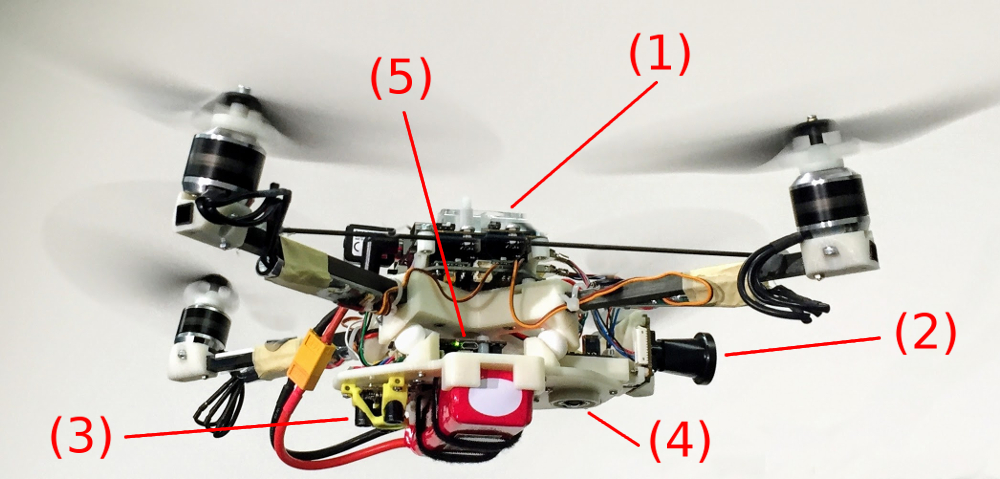}
  \caption{The quadrotor platform used in the experiments. 
	(1) Onboard computer.
	(2) Forward-facing fisheye camera. 
	(3) TeraRanger One distance sensor and (4) downward-facing camera, both used \emph{solely during the recovery phase}. 
	(5) PX4 autopilot.
  The motors are tilted by \SI{15}{\degree} to provide three times more yaw-control action, while only losing \SI{3}{\percent} of the collective thrust.}
  \label{fig:quadrotor_platform}
\end{figure}

We tested the proposed framework on a custom-made quadrotor, assembled from off-the-shelf hardware, 3D printed parts, and self-designed electronic components (see Fig.~\ref{fig:quadrotor_platform}).
The frame of the vehicle is composed of a 3D printed center cross and four carbon fiber profiles as arms.
Actuation is guaranteed by four RCTimer MT2830 motors, controlled by Afro Slim ESC speed controllers.
The motors are tilted by \SI{15}{\degree} to provide three times more yaw-control action, while only losing \SI{3}{\percent} of the collective thrust.

Our quadrotor is equipped with a PX4FMU autopilot that contains an IMU and a micro controller on which our custom low-level controller runs.
Trajectory planning, state estimation and high-level control run on an Odroid-XU4 single-board computer.
Our algorithms have been implemented in ROS, running on Ubuntu 14.04.
Communication between the Odroid and the PX4 runs over {\it UART}.

Gap-detection is done through a forward-facing fisheye camera (MatrixVision mvBlueFOX-MLC200w $752 \times 480$-pixel monochrome camera with a \SI{180}{\degree} lens), which ensures that the gap can be tracked until very close.
To allow the robot to execute the recovery maneuver after traversing the gap, we mounted the same hardware detailed in~\cite{Faessler15icra},
which consists of a TeraRanger One distance sensor and a downward-facing camera. Notice, however, that these are \emph{not used} for state estimation before passing the gap but \emph{only} 
to recover and switch into stable hovering after the traverse.

\begin{figure}[tb]
\centering
\includegraphics[width=0.6\linewidth]{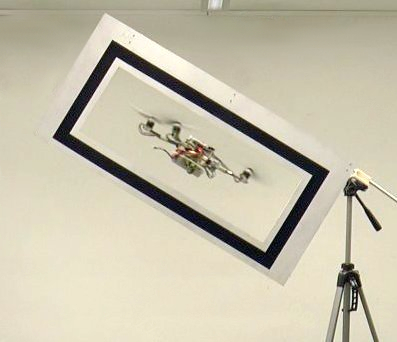}
\caption{Our quadrotor during a traverse.}
\label{fig:quadingap}
\end{figure}

The overall weight of the vehicle is \SI{830}{\gram}, while its dimension are \SI[product-units = single]{55 x 12}{\centi \meter} (largest length measured between propeller tips).
The dimensions of the rectangular gap are \SI[product-units = single]{80 x 28}{\centi \meter}.
When the vehicle is at the center of the gap, the tolerances along the long side and short sides are only \SI{12.5}{\centi \meter}, 
and \SI{8}{\centi \meter}, respectively (cf. Fig.~\ref{fig:quadingap}). This highlights that the traverse trajectory must be followed with centimeter accuracy to avoid a collision.

The parameters of the traverse trajectory (Sec.~\ref{sec:traverse_optimization}) have been set as
$\mathrm{v}_{\mathrm{0,max}} =  \SI{3}{\meter \per \second}$, $\mathrm{d}_{\mathrm{min}} =  \SI{0.25}{\centi \meter}$.
The normalization constants $\bar{\theta}$ and $\bar{d}$, introduced in Sec.~\ref{sec:choice_of_trajectory}, have been manually tuned to let the quadrotor start the maneuver close enough to render vision-based pose estimation reliable and, at the same time, keep the gap as close as possible to the center of the image.

The dynamic model and the control algorithm used in this work are the same presented in \cite{Faessler15icra}. 
We refer the reader to that for further details.

\subsection{Results} \label{sec:experiments_results}

To demonstrate the effectiveness of the proposed method, we flew our quadrotor through a gap inclined at different orientations. 
We consider both rotations around the world $x$ and $y$ axes, and denote them as \emph{roll} and \emph{pitch}, respectively.
Overall, we ran 35 experiments with the roll angle ranging between \SI{0}{\degree} and \SI{45}{\degree} and the pitch angle between \SI{0}{\degree} and \SI{30}{\degree}.
We discuss the choice of these values in Sec.~\ref{sec:gap_configuration}.
With the gap inclined at \SI{45}{\degree}, the quadrotor reaches speeds of \SI{3}{\meter \per \second} and angular velocities of \SI{400}{\degree \per \second}. 

We define an experiment as successful if the quadrotor passes through the gap without collision and recovers and locks to a hover position. 
We achieved a remarkable success rate of $80\%$.
When failure occurred, we found this to be caused by a persistent absence of a pose estimate from the gap detector during the approach trajectory. 
This led to a large error in matching the initial conditions of the traverse trajectory, which resulted in a collision with the frame of the gap.

\begin{figure*}
  \centering
  \subfloat[Gap: \SI{20}{\degree} roll, \SI{0}{\degree} pitch.]{\includegraphics[width=0.25\linewidth]{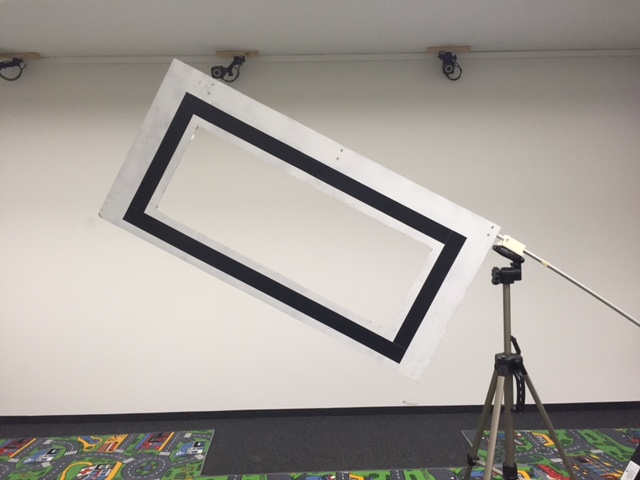}\label{fig:gap20}} \ \ \ \ \ \ \ \ \ \
  \subfloat[Gap: \SI{45}{\degree} roll, \SI{0}{\degree} pitch.]{\includegraphics[width=0.25\linewidth]{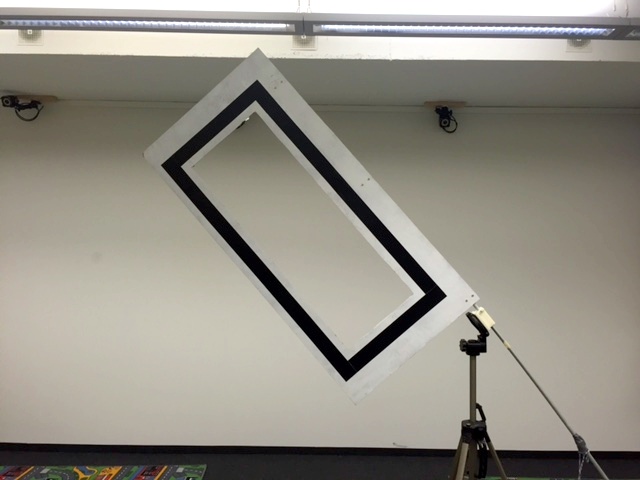}\label{fig:gap45}} \ \ \ \ \ \ \ \ \ \
  \subfloat[Gap: \SI{30}{\degree} roll, \SI{30}{\degree} pitch.]{\includegraphics[width=0.25\linewidth]{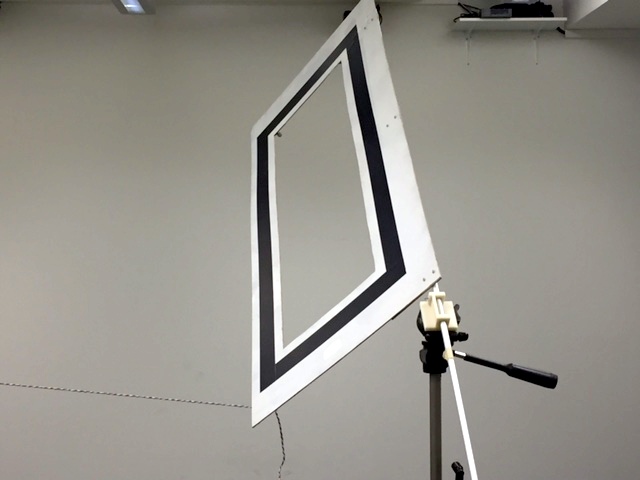}\label{fig:gap30-30-side}} \hfill
  \subfloat[Gap: \SI{20}{\degree} roll, \SI{0}{\degree} pitch.]{\includegraphics[width=0.33\linewidth]{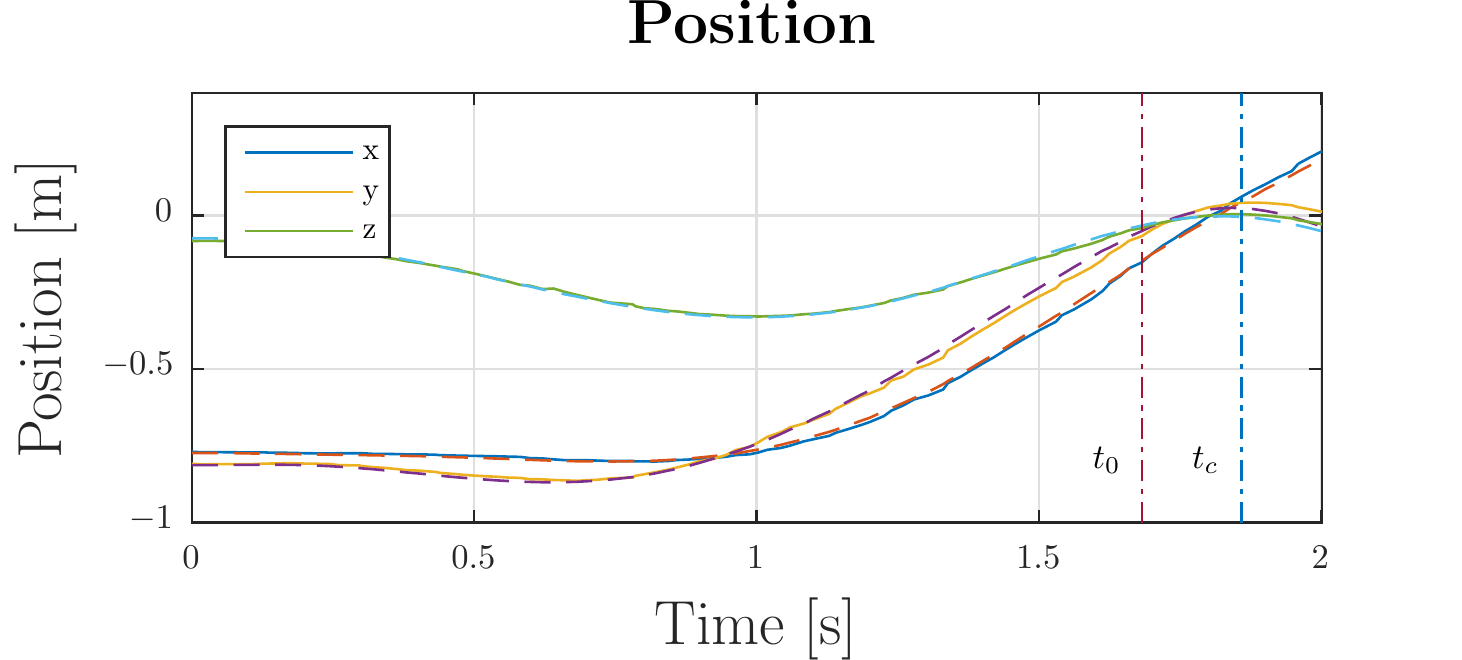}\label{fig:exp1_pos}}
  \subfloat[Gap: \SI{45}{\degree} roll, \SI{0}{\degree} pitch.]{\includegraphics[width=0.33\linewidth]{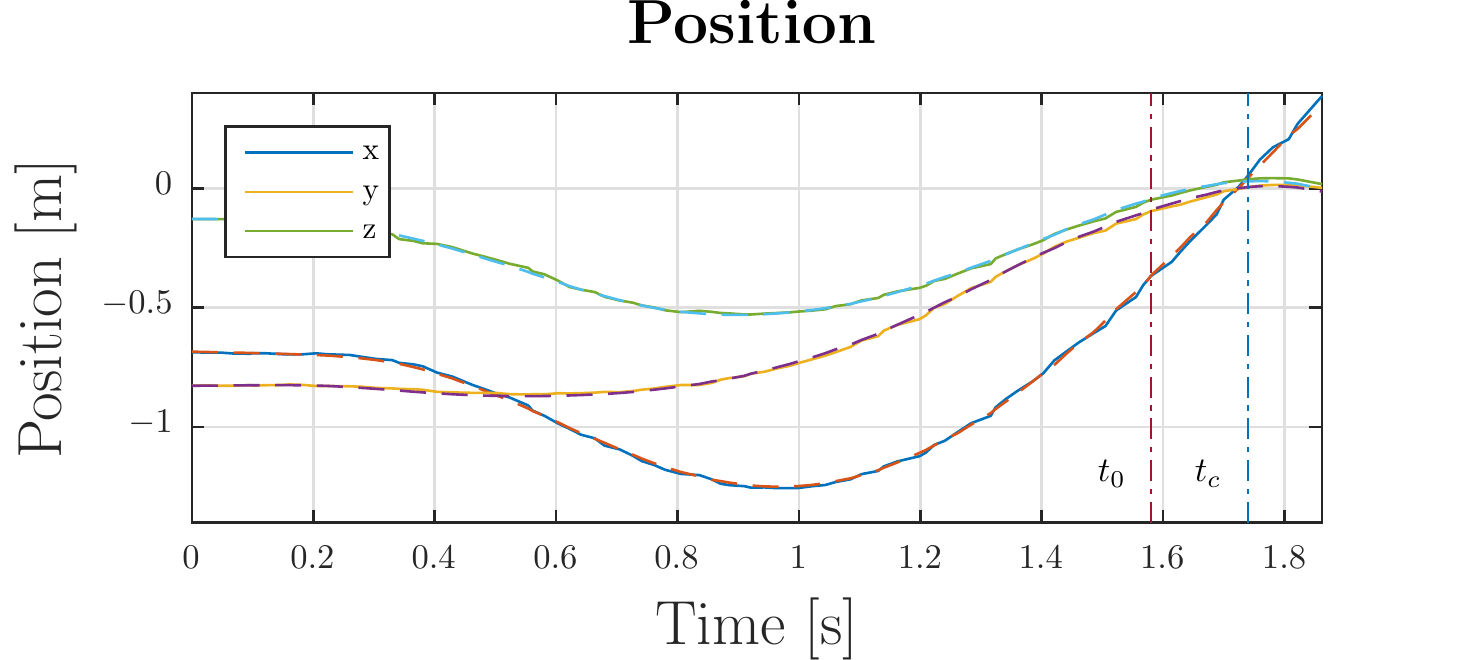}\label{fig:exp2_pos}}
  \subfloat[Gap: \SI{30}{\degree} roll, \SI{30}{\degree} pitch.]{\includegraphics[width=0.33\linewidth]{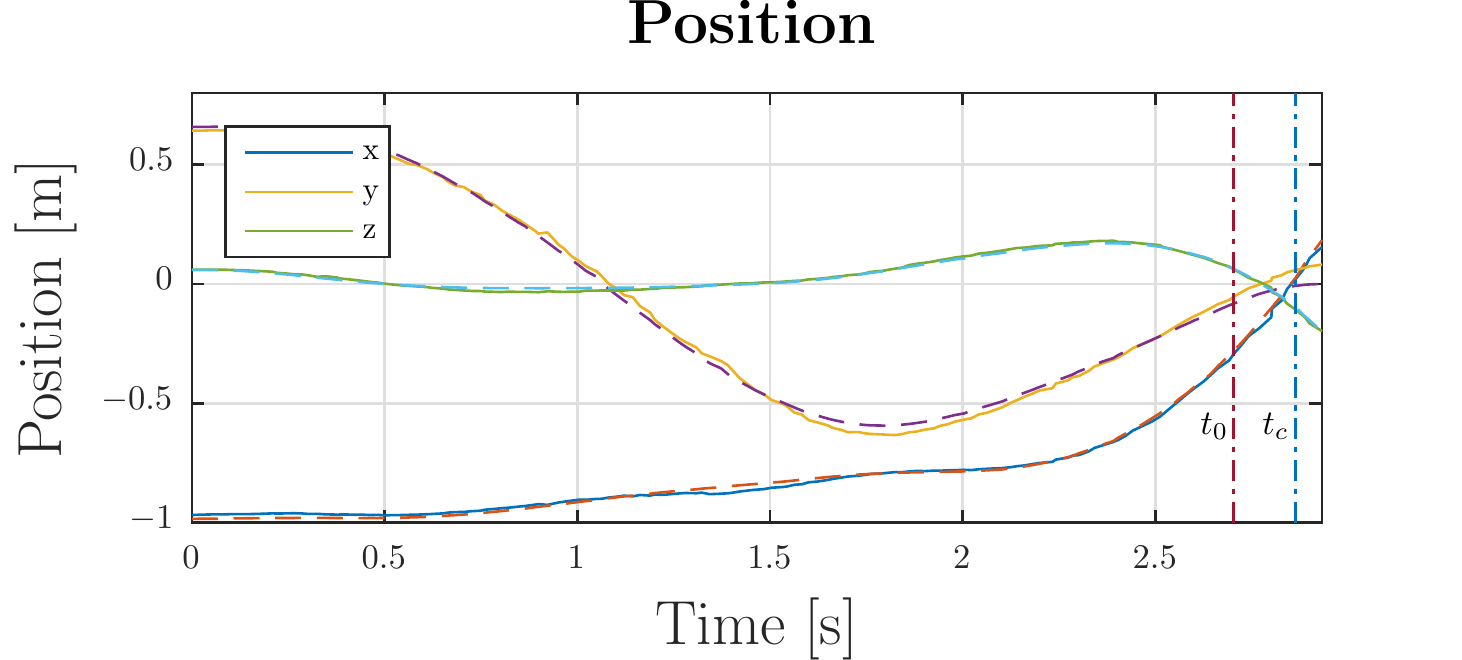}\label{fig:exp3_pos}} \hfill
  \subfloat[Gap: \SI{20}{\degree} roll, \SI{0}{\degree} pitch.]{\includegraphics[width=0.33\linewidth]{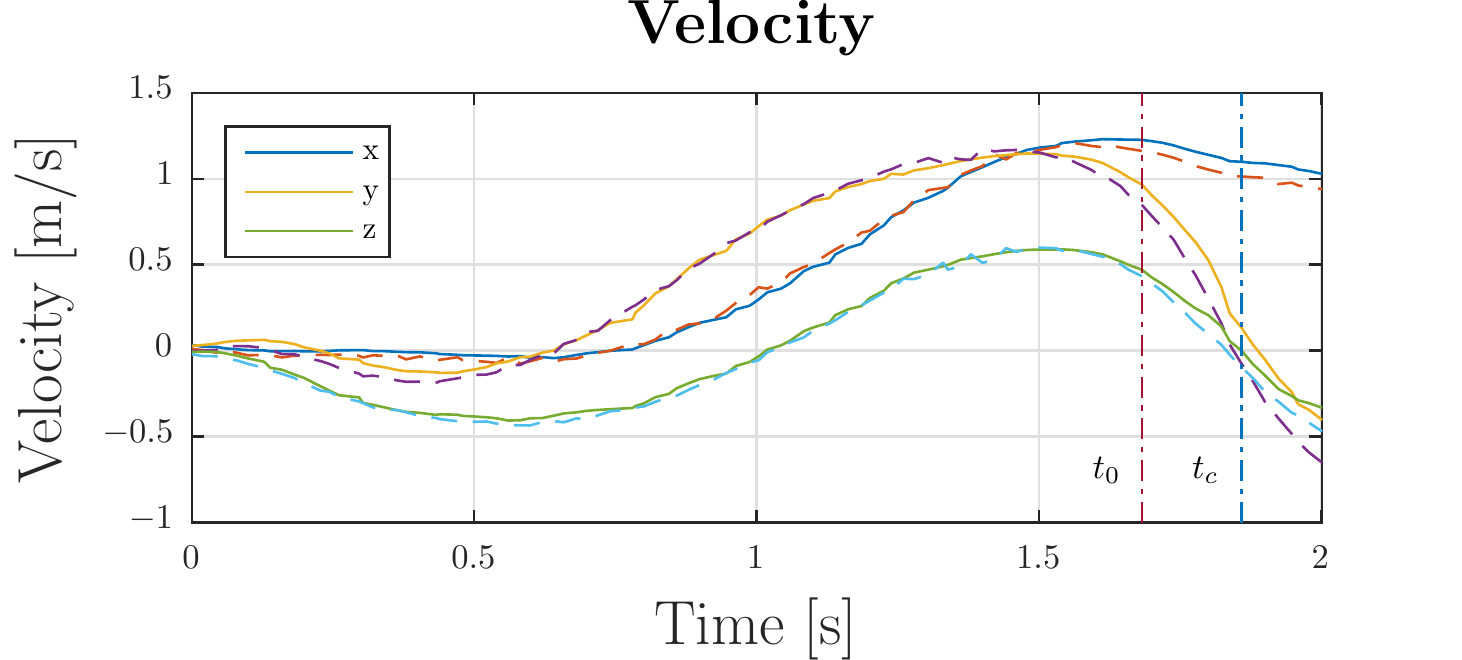}\label{fig:exp1_vel}}
  \subfloat[Gap: \SI{45}{\degree} roll, \SI{0}{\degree} pitch.]{\includegraphics[width=0.33\linewidth]{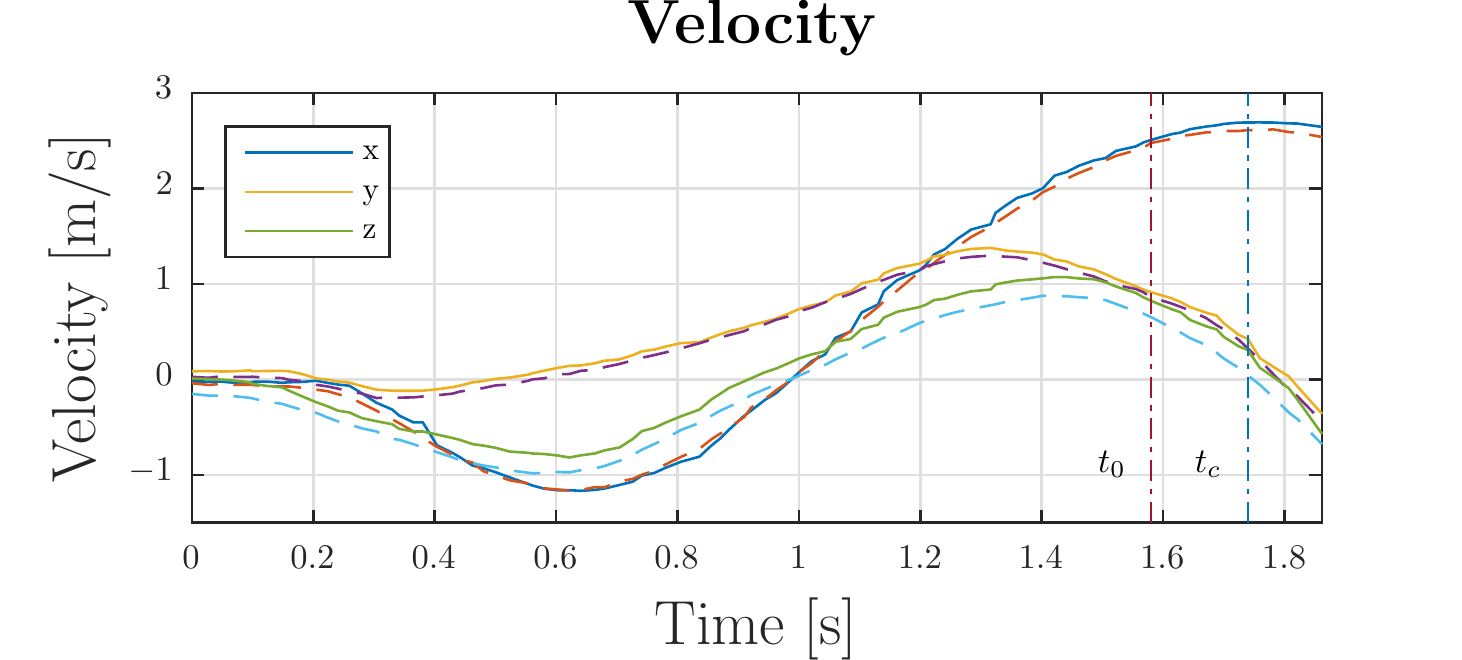}\label{fig:exp2_vel}}
  \subfloat[Gap: \SI{30}{\degree} roll, \SI{30}{\degree} pitch.]{\includegraphics[width=0.33\linewidth]{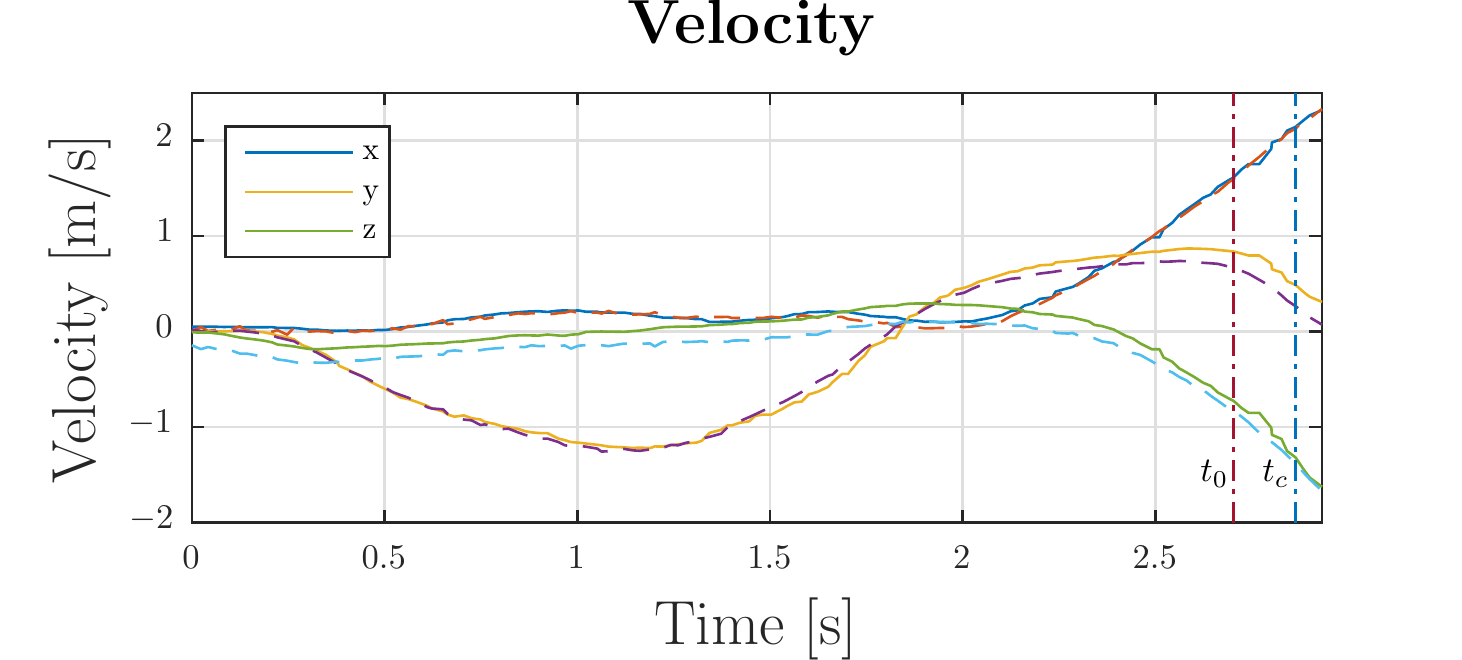}\label{fig:exp3_vel}} \hfill
  \subfloat[Gap: \SI{20}{\degree} roll, \SI{0}{\degree} pitch.]{\includegraphics[width=0.33\linewidth]{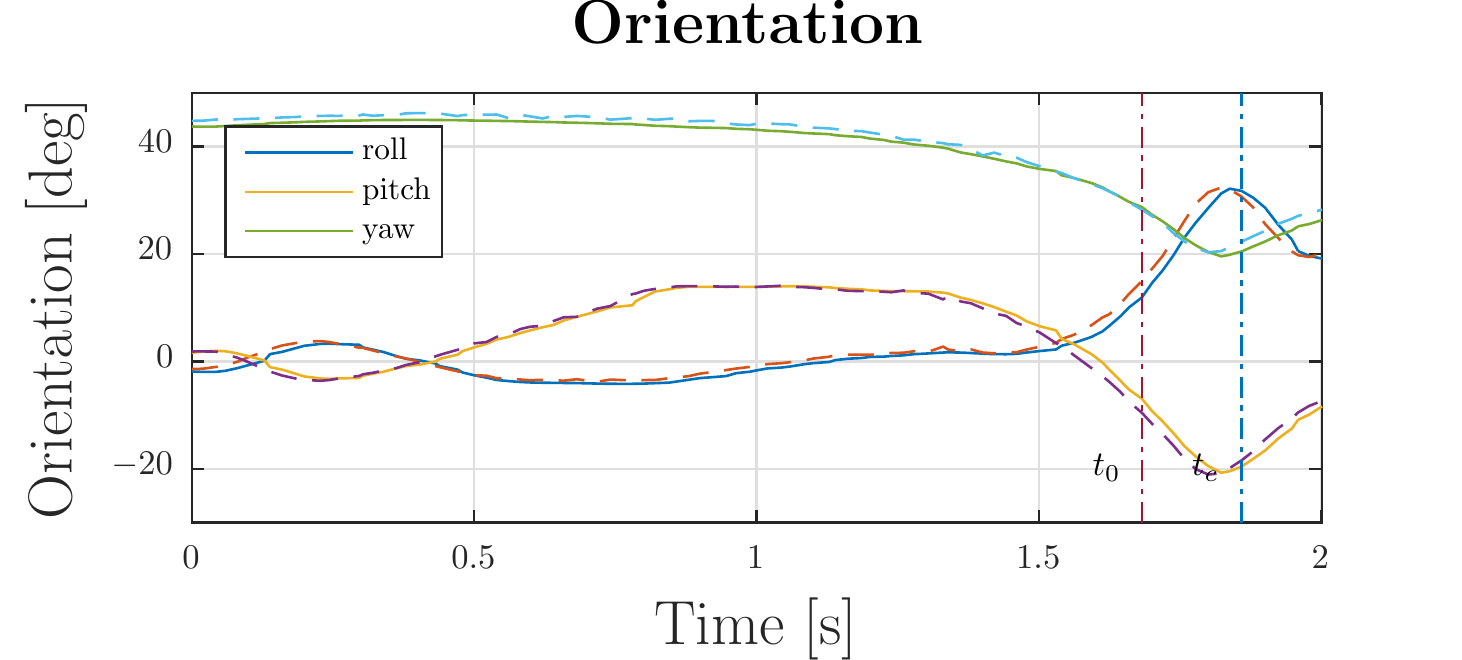}\label{fig:exp1_orient}}
  \subfloat[Gap: \SI{45}{\degree} roll, \SI{0}{\degree} pitch.]{\includegraphics[width=0.33\linewidth]{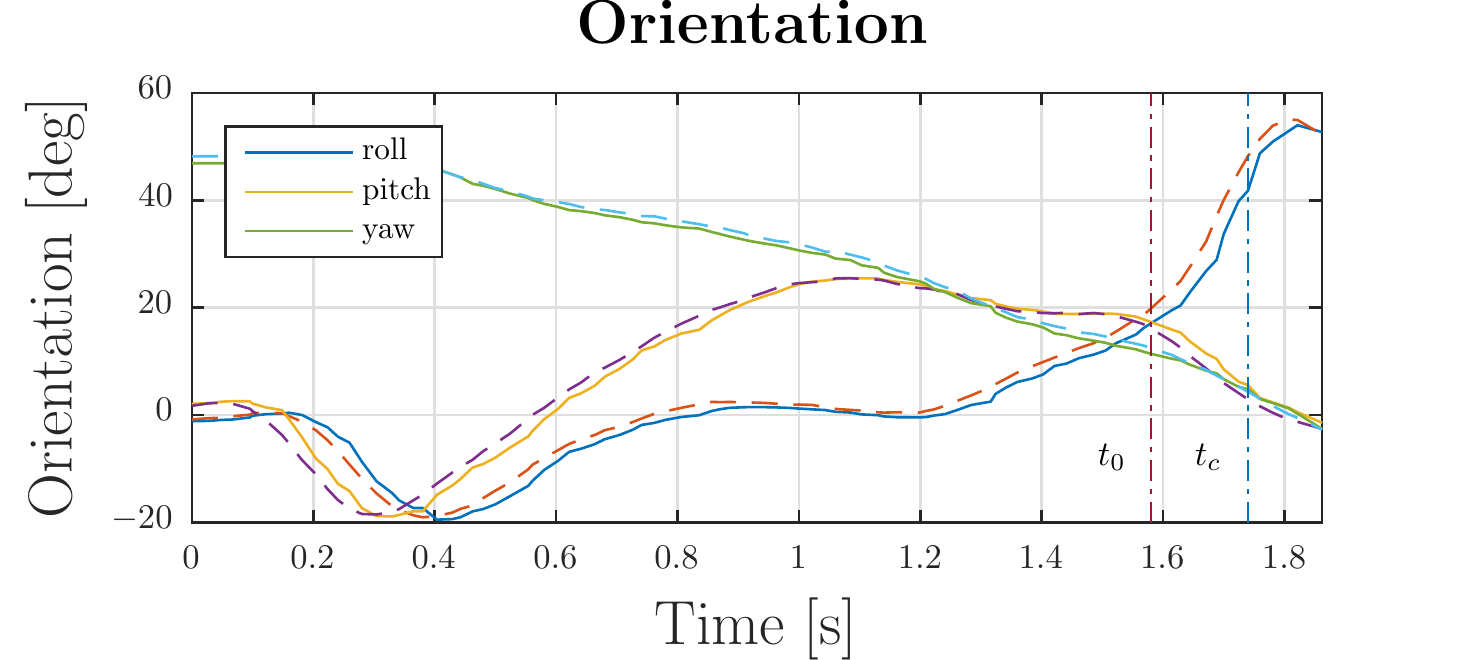}\label{fig:exp2_orient}}
  \subfloat[Gap: \SI{30}{\degree} roll, \SI{30}{\degree} pitch.]{\includegraphics[width=0.33\linewidth]{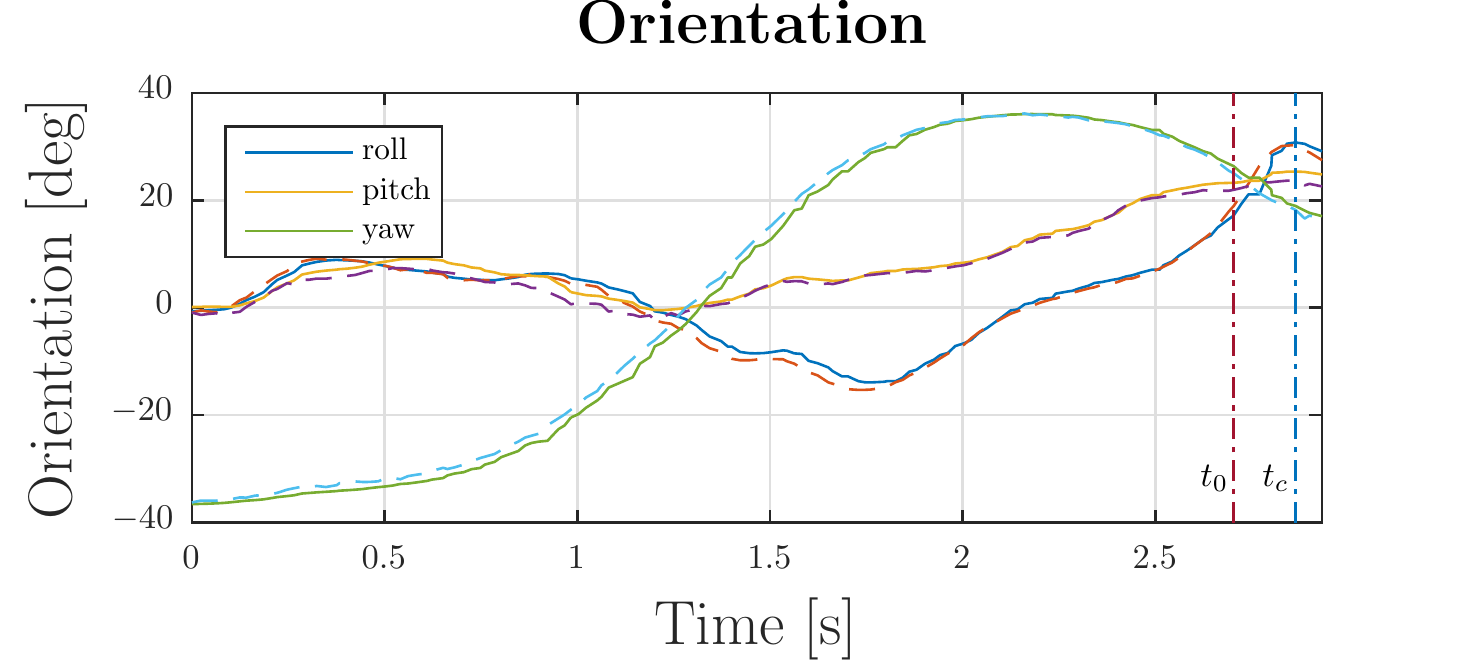}\label{fig:exp3_orient}} 
  
  \caption{Comparison between ground truth and estimated position (top), velocity (center), and orientation (bottom). 
					 Each column depicts the result of an experiment conducted with a different configuration of the gap:
						\protect\subref{fig:exp1_pos}, \protect\subref{fig:exp1_vel} and \protect\subref{fig:exp1_orient} \SI{20}{\degree} of roll and \SI{0}{\degree} of pitch; 
						\protect\subref{fig:exp2_pos}, \protect\subref{fig:exp2_vel} and \protect\subref{fig:exp2_orient} \SI{45}{\degree} of roll and \SI{0}{\degree} of pitch; 
						\protect\subref{fig:exp3_pos}, \protect\subref{fig:exp3_vel} and \protect\subref{fig:exp3_orient} \SI{30}{\degree} of roll and \SI{30}{\degree} of pitch.
						The approach trajectory starts at $\t = 0$ and ends at $\t = \t_0$, when the traverse trajectory is executed. 
						The quadrotor reaches the center of the gap at $\t = \t_c$ and starts the recovery maneuver at the final time of each plot.
						We refer the reader to the accompanying video for further experiments with different orientations of the gap.}
  \label{fig:experiments_plots}
\end{figure*}

\begin{figure}[tb]
\centering
\includegraphics[width=\linewidth]{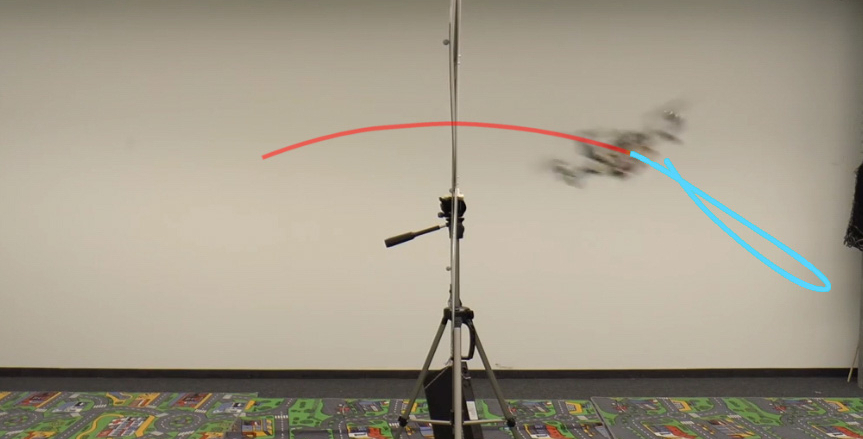}
\caption{Our quadrotor executing the whole trajectory split into approach (blue), traverse (red).}
\label{fig:trajectory_on_screenshot}
\end{figure}

Figure~\ref{fig:experiments_plots} shows the estimated position, velocity, and orientation against ground truth for some of the most significant experiments and for different orientations of the gap 
(namely: \SI{20}{\degree} roll, \SI{0}{\degree} pitch; \SI{45}{\degree} roll, \SI{0}{\degree} pitch; and \SI{30}{\degree} roll, \SI{30}{\degree} pitch).
Ground truth is recorded from an OptiTrack motion-capture system.
It can be observed that the desired trajectories were tracked remarkably well.
Table~\ref{table:error_gap_center_statistics} reports the statistics of the errors when the quadrotor passes through the plane in which the gap lies (i.e., at $t = t_c$), measured as the distance between actual and desired state.
These statistics include both the successful and the unsuccessful experiments.
The average of the norm of the position error at the center of the gap was \SI{0.06}{\meter}, with a standard deviation of \SI{0.05}{\meter}.
The average of the norm of the velocity error was below \SI{0.19}{\meter \per \second}, with a standard deviation of \SI{0.20}{\meter \per \second}.
We refer the reader to the attached video for further experiments with different orientations of the gap. 
Figure~\ref{fig:trajectory_on_screenshot} shows a picture of one of the experiments with the executed approach and traverse trajectories marked in color.

\begin{table}[]
  \centering
  \begin{tabular}{r|ccc|ccc|cc}
    \toprule &
		\multicolumn{3}{c|}{\textbf{Position [\SI{}{\meter}]}} &
		\multicolumn{3}{c|}{\textbf{Velocity [\SI{}{\meter \per \second}]}} &
		\multicolumn{2}{c}{\textbf{Orientation [\SI{}{\degree}]}} \\
	\midrule
		& \textbf{x} & \textbf{y} & \textbf{z} &
		  \textbf{x} & \textbf{y} & \textbf{z} &
		  \textbf{roll} & \textbf{pitch} \\
	\midrule
		$\mu$    & 0.04 & 0.04 & 0.03 & 0.09 & 0.15 & 0.08 & 6.04 & 8.89 \\
	\midrule
	    $\sigma$ & 0.03 & 0.02 & 0.03 & 0.08 & 0.10 & 0.06 & 3.70 & 5.85 \\
	    \bottomrule
  \end{tabular}
    \caption{Position, velocity and orientation error statistics at time $\t = \t_c$. 
  The mean error $\mu$ and the standard deviation $\sigma$ are computed using ground truth data gathered from 35 experiments conducted with the gap at different orientations.
	}
  \label{table:error_gap_center_statistics}
\end{table}

%% file: chapters/discussion.tex
\section{Discussion} \label{sec:discussion}

In this section, we discuss our approach and provide more insights into our experiments.

\subsection{Replanning}
The method we use to compute the approach meneuver~\cite{Mueller15tro} can fail to verify whether a trajectory is feasible or not, as also highlighted by the authors.
This usually happens when the time duration of the trajectory is short.
In such a case, we skip the replanning and provide the last available approach trajectory to our controller.

\subsection{Trajectory Computation Times}
The trajectory planning approach we adopt for the approach phase is fast enough to compute and test $40,000$ trajectories in less then one second, even with the additional computational load induced by our check on the gap perception.
The computation of each trajectory on the on-board computer takes on average \SI[separate-uncertainty=true]{0.240 \pm 0.106}{\milli\second}, including: (i) generation of the trajectory; (ii) feasibility check; (iii) trajectory sampling and computation of the yaw angle for each sample; (iv) evaluation of the cost function described in~\eqref{eq:approach_cost_function}; (v) comparison with the current best candidate.
It is important to point out that these values do not apply to the \emph{replanning} of the approach trajectory during its execution, since the initial state is constrained by the current state of the vehicle and there is no cost function to evaluate.
In such a case, the computation is much faster and for each trajectory it only takes \SI[separate-uncertainty=true]{0.018 \pm 0.011}{\milli\second} on average.

\subsection{Gap configuration} \label{sec:gap_configuration}
Our trajectory generation formulation is able to provide feasible trajectories with any configuration of the gap, 
e.g., when the gap is perfectly vertical (\SI{90}{\degree} roll angle) or perfectly horizontal (\SI{90}{\degree} pitch angle). 
However, in our experiments we limit the roll angle of the gap between \SI{0}{\degree} and \SI{45}{\degree} and the pitch angle between \SI{0}{\degree} and \SI{30}{\degree}.
We do this for two reasons.
First, when the gap is heavily pitched, the quadrotor needs more space to reach the initial conditions of the traverse from hover.
This renders the gap barely or not visible at the start of the approach, increasing the uncertainty in the pose estimation.
Second, extreme configurations, such as roll angles of the gap up to \SI{90}{\degree}, require high angular velocities in order to let the quadrotor align its orientation with that of the gap.
This makes gap detection difficult, if not impossible, due to motion blur.
Also, our current experimental setup does not allow us to apply the torques necessary to reach high angular velocities because of the inertia of the platform and motor saturations.

\subsection{Dealing with Missing Gap Detections}
The algorithm proposed in Sec.~\ref{sec:window_tracking} fuses the poses from gap detection with IMU readings to provide the full state estimate during the approach maneuver. 
In case of motion blur, due to high angular velocities, or when the vehicle is too close to the gap, the gap detection algorithm does not return any pose estimate.
However, these situations do not represent an issue during short periods of time (a few tenths of a second). In these cases, the state estimate from the sensor fusion module is still available and reliable through the IMU.

%% file: chapters/conclusion.tex
\section{Conclusion} \label{sec:conclusion}
We developed a system that lets a quadrotor vehicle safely pass through a narrow inclined gap using only onboard sensing and computing.
Full state estimation is provided by fusing gap detections from a forward-facing onboard camera and an IMU.

To tackle the problems arising from the varying uncertainty from the vision-based state estimation, we coupled perception and control by computing trajectories that facilitate state estimation by always keeping the gap in the image of the onboard camera.

We successfully evaluated and demonstrated the approach in many real-world experiments.
To the best of our knowledge, this is the first work that addresses and achieves autonomous, aggressive flight through narrow gaps using only onboard sensing and computing, and without requiring prior knowledge of the pose of the gap.
We believe that this is a major step forward autonomous quadrotor flight in complex environments with onboard sensing and computing.